# Ethylene Leak Detection Based on Infrared Imaging: A Benchmark

Xuanchao Ma
Faculty of Information Technology
Beijing University of Technology
Beijing, China
maxuanchao@emails.bjut.edu.cn

Yuchen Liu
Faculty of Information Technology
Beijing University of Technology
Beijing, China
kuailexiaoliu@emails.bjut.edu.cn

*Abstract*—Ethylene leakage detection has become one of the most important research directions in the field of target detection due to the fact that ethylene leakage in the petrochemical industry is closely related to production safety and environmental pollution. Under infrared conditions, there are many factors that affect the texture characteristics of ethylene, such as ethylene concentration, background, and so on. We find that the detection criteria used in infrared imaging ethylene leakage detection research cannot fully reflect real-world production conditions, which is not conducive to evaluate the performance of current image-based target detection methods. Therefore, we create a new infrared image dataset of ethylene leakage with different concentrations and backgrounds, including 54275 images. We use the proposed dataset benchmark to evaluate seven advanced image-based target detection algorithms. Experimental results demonstrate the performance and limitations of existing algorithms, and the dataset benchmark has good versatility and effectiveness.

## I. INTRODUCTION

Ethylene industry is the core of the petrochemical industry and occupies an important position in the national economy. Ethylene production is regarded internationally as an important indicator of a country's petrochemical development level. With the continuous improvement of large-scale automation devices in industrial production such as petrochemical enterprises, ethylene leakage often occurs during production. If the leaked ethylene is not handled in a timely manner, it will pose a serious threat to safety in production and the ecological environment. Therefore, ethylene leakage detection has long been a concern. Traditional ethylene detection mainly uses sensors to detect ethylene [1]-[3]. Common types of sensors include semiconductor oxide sensors, electrochemical sensors, and photoionization sensors. However, there are obvious shortcomings in traditional sensor detection methods for ethylene leakage detection in production units of petrochemical enterprises. Firstly, traditional sensor detection methods belong to point measurement, and their contact principle makes many areas to be measured inaccessible, which is unable to adapt to large-scale dynamic real-time detection [4]-[7]. Second, ethylene diffusion is highly susceptible to weather and other factors, resulting in low detection accuracy and generally unrecognizable leakage by human eyes. It is impossible to determine the leak point simply by reading data from the device. Therefore, traditional ethylene detection methods mainly rely on sensors and manual judgment. Due to its unique characteristics, using this method cannot detect ethylene leakage in real time and accurately. With the development of technology such as machine vision, image-based technology provides new solutions for ethylene detection tasks.

Early research has shown that natural images captured by cameras have multiple statistical laws [8]-[13], which have been successfully used for image restoration, quality prediction, and so on. In recent years, researchers have widely applied image-based technology to different target detection fields. In order to reduce pollution, image-based particle detection methods [14]-[19] and smoke detection methods [20]-[27] have been widely studied topics. The existing image-based ethylene detection methods have the characteristics of rapid response, non-contact, and large monitoring area [28]. However, as far as we know, the existing infrared image-based ethylene dataset benchmark cannot fully consider the factors that affect the texture characteristics of the ethylene, such as the concentration and background of the ethylene, which cannot fully reflect the production situation in the real world. This makes it impossible to accurately evaluate the performance of existing advanced image-based target detection methods. In addition, actual industrial control systems can only make basic decisions based on limited ethylene leakage detection results, and new developed algorithms using these datasets cannot provide additional information to help control systems generate more accurate regulation. Therefore, creating a new infrared image dataset benchmark for ethylene leakage detection with different concentrations and backgrounds has important practical implications for solving the above engineering problems. In this paper, we propose and establish a ethylene leakage detection dataset benchmark based on infrared imaging, including 54275 images.

The remaining structure of this article is organized as follows. The process of creating a dataset is described in depth in Section 2. Section 3 examines the effectiveness of the dataset and the performance of the algorithm. The main conclusions are presented in Section 4.

## II. DATASET BENCHMARK SETTINGS

Under infrared conditions, there are many factors that affect the gray value of ethylene, such as its concentration and background. However, the texture characteristics of ethylene vary greatly depending on different concentrations and backgrounds. Therefore, we have created a new infrared image dataset for ethylene leakage detection, containing different concentrations and backgrounds, to accurately evaluate the performance of existing advanced image-based target detection methods.

To facilitate the three stages of algorithm development, optimization, and evaluation, we divide the ethylene leakage infrared image dataset into three parts: a training set, a validation set, and a test set. The training set is used to develop training algorithms, while the validation set is used to optimize algorithms by adjusting parameters, network structures, and selecting features. Finally, the test set is utilized to evaluate the performance of the algorithms. We organize the images in the dataset into two categories: with and without ethylene leakage. The ethylene leakage infrared videos used in the constructed dataset are captured using a specific infrared camera with a frame rate of 25 frames per second. We obtain one frame for every two frames during a time interval from various types of videos. The image block area that meet the characteristic requirements for the ethylene leakage location is identified through block processing of the infrared images, and the resulting images are stored in PNG format. To ensure that the dataset contained ethylene leakage images of different concentrations, we make appropriate adjustments to the ethylene leakage speed and capture video images of both early and late ethylene leakage situations. Through observation of video images, it can be found that ethylene has a significant diffusivity at the initial stage of leakage, which is accompanied by a dynamic diffusion process from nothing to something, from small to large targets. The edge contour of ethylene often shows a gradual change from the center of the ethylene to the edge region, with a transition from thick to light, which is evident in the infrared images, as the gray value from the center region of the ethylene to the edge region changes from large to small or from small to large. To meet the conditions for including different ethylene leakage image backgrounds in the dataset, we have set up indoor and outdoor video image shooting environments for both cases with and without ethylene leakage. The image background in indoor environments is relatively simple, and the background of ethylene leakage is mostly a single wall background. The image background in the outdoor environment is relatively complex, and the drift direction of ethylene leakage is easily affected by the outdoor environment, increasing the diversity of ethylene leakage images.

To classify images based on the presence or absence of ethylene leakage, a subjective scoring method is utilized. To minimize the learning error caused by inaccurate labels, the label of an image should be derived from multiple labeling outcomes. In this study, 25 people are involved in labeling the dataset during the establishment process. The process of selecting the final label from multiple results can be compared to the hard voting process in integrated learning. To be specific, let $X = [x_1, x_2, ..., x_n]$, that $n$ is the number of categories and $x_i$ is the number of votes for an image has received to be classified in a certain category ($n$ equals two in this paper). The sum of $x_i$ should also equal M, where M is the total number of votes cast for an image, which is 25. The following is the image's final category:

$$x = argrmax(X) \quad (1)$$

$argrmax$ denotes that the final category of the image is obtained from the subscript $i$ that maximizes the value of X.

In order to simulate the shooting effect of the camera under harsh conditions, we adopt a noise injection strategy to establish a more realistic and challenging ethylene leakage image dataset. On the one hand, noise injection can easily increase the number of ethylene leakage images. On the other hand, noise is inevitably introduced during the image acquisition process, which is closer to the real situation. This strategy adds random noise to the images in the training, validation, and test sets, effectively making the dataset closer to actual data collection scenarios. By adding random noise to the images in the training set, verification set, and test set, we have achieved an increase in the number of images in the dataset, which is more conducive to model training.

The training set and validation set each include 20000 images, of which both datasets include 10000 images with ethylene leakage and 10000 images without ethylene leakage. There are 14275 images in the test set, including 9000 images with ethylene leakage and 5725 images without ethylene leakage. Some images in the dataset are shown in Fig. 1:

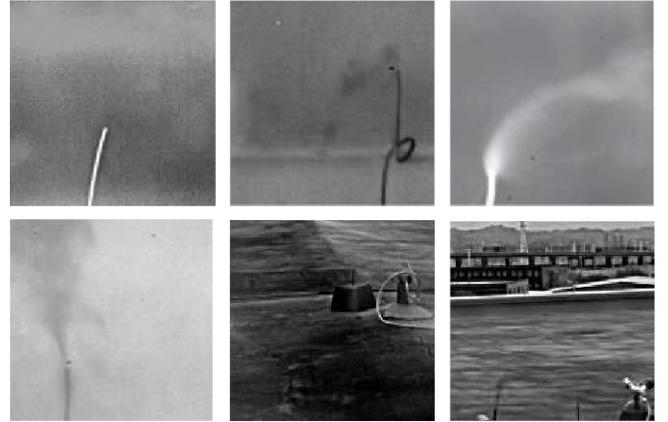

Figure 1. Image example of ethylene leakage dataset based on infrared imaging.

III. EXPERIMENT

In this section, we evaluate the performance of multiple image-based target detection algorithms using the proposed benchmark, and assess the effectiveness and universality of the benchmark dataset. This section consists of the following four aspects: operating environment, evaluation indicators, competing methods, and performance comparison.

*A. Operating environment*

All models used in this paper have the same configuration to allow for a fair comparison of model performance. Our experimental framework is PyTorch. An Ubuntu computer running the Inter(R) Gold 6248R CPU at 3.00GHz and an NVIDIA GeForce RTX 3090 graphics card power the experimental environment. Table I displays the experiment's hyperparameter setting.

TABLE I. HYPERPARAMETER SETTINGS USED IN THE EXPERIMENT

|  | **Hyperparameter settings** |
|---|---|
| Epoch | 300 |
| Batch Size | 128 |
| Optimizer | SGD base lr:0.001 Momentum:0.9 Weight_decay:1E-5 |
| Loss | CrossEntropyLoss |
| Lr Scheduler | Learning rate scales linearly from base_lr to 1E-5 |

*B. Evaluation indicators*

To quantify the performance of each model, we leverage three typically used evaluation indicators, including accuracy rate (AR), detection rate (DR), and false alarm rate (FAR). From these indicators, we can analyze the performance of the algorithm and evaluate the effectiveness and universality of the dataset. The following are the definitions of these indicators: the AR is defined as the proportion of the correct results obtained by the model used in leak and non-leak detection to the total number of samples

$$AR = \frac{P_1 + N_2}{T_1 + T_2} \times 100\% \quad (2)$$

where $T_1$ and $T_2$ are the number of positive samples and the number of negative samples, respectively; $P_1$ is the number of correctly classified true positive samples; $N_2$ is the number of correctly classified true negative samples. The DR is defined as the proportion of correct results to the total number of positive samples

$$DR = \frac{P_1}{T_1} \times 100\% \quad (3)$$

and the FAR is defined as

$$FAR = \frac{N_1}{T_2} \times 100\% \quad (4)$$

where $N_1$ is the number of negative samples incorrectly classified as positive samples.

Based on the formulas defined above, an excellent classification model is expected to achieve large values in AR and DR, while a small value in FAR. The experimental results can provide calculated values of these indicators, which can evaluate the performance of the model. A great model should strive to achieve greater values on these indicators as far as possible.

*C. Competing methods*

The comparative experiments are conducted using established dataset benchmark. We use seven advanced algorithm models, including ResNet-18 [29], DenseNet-121 [30], ViT [31], LocalViT [32], PiT [33], MobileViT [34] and SBasicViT [35]. In recent years, seven types of networks have made significant progress in general image classification datasets, with great accuracy and fast computing speed.

*D. Performance comparison*

Based on the established dataset benchmark, we use three typical evaluation indicators to record and compare the performance of seven advanced models: AR, DR, and FAR. Through analysis of these indicators, we can not only evaluate the validity and universality of the dataset, but also further quantify the performance and limitations of the algorithm to improve the performance of the model in classification tasks. The performance results of the seven advanced networks are shown in Table II.

TABLE II. PERFORMANCE COMPARISON OF SEVEN ADVANCE MODELS

| **Situation** | **AR** | **DR** | **FAR** |
|---|---|---|---|
| ResNet-18 [29] | 78.6%±2.1% | 70.1%±3.0% | 17.6%±1.8% |
| DenseNet-121 [30] | 79.7%±5.0% | 79.5%±7.7% | 12.0%±4.7% |
| ViT [31] | 73.2%±4.7% | 60.6%±7.7% | 23.1%±4.7% |
| LocalViT [32] | 77.8%±2.1% | 71.2%±3.2% | 16.9%±1.9% |
| PiT [33] | 73.6%±4.4% | 64.2%±5.3% | 20.1%±9.4% |
| MobileViT [34] | 77.8%±2.3% | 69.6%±5.7% | 17.8%±3.4% |
| SBasicViT [35] | 73.1%±2.0% | 65.8%±2.5% | 20.1%±1.5% |

An excellent classification model is expected to achieve large values in AR and DR, while a small value in FAR. Table II shows that DenseNet-121 [30] performs best, with an accuracy rate of about 79.7%, a detection rate of about 79.5%, and a false alarm rate of about 12.0%. In other words, in binary classification testing, it can effectively detect whether there is ethylene leakage in the image. In addition, we find that the accuracy and detection rates of ViT [31], PiT [33] and SBasicViT [35] are significantly lower than those of the other four models, while the false alarm rate is high, which indicates that their generalization performance may be slightly worse and the performance of the model needs to be improved. In real-world applications, the models must accurately identify specific scenarios without confusion, requiring improved accuracy while maintaining detection rate.

There has been a lot of work put into building the database to prevent importing inaccurate label information as much as possible. This demonstrates that there are additional causes for the uncertainty, like backdrop confusion in the image or color confusion brought on by changes in illumination. Features that are unaffected by environmental noise and brightness variations should be further mined by the models. Our experimental results have successfully demonstrated the performance and limitations of existing advanced algorithms, and the dataset benchmark is effective and universal, which can provide a basic platform for image-based target detection algorithm research.

## IV. CONCLUSION

Ethylene leakage detection is crucial for reducing air pollution and ensuring safe production. In this article, we have created a ethylene leakage dataset benchmark based on infrared images specifically for ethylene detection. Firstly, we use a subjective scoring method to label the images with and without ethylene, and then apply a noise injection strategy to simulate the effects of poor camera conditions and increase the size of the dataset. The final dataset contains 54275 images. Furthermore, we evaluate the effectiveness and integrity of the dataset by using it to test the performance and limitations of seven advanced image-based target detection algorithms. The experimental results show that the dataset has great universality and effectiveness, which can provide a basic platform for the research of image-based target detection algorithms. In our future work, we will strive to mine ethylene leakage characteristics using self-supervision and other technologies to improve the performance of fully supervised tasks and solve the challenge of detecting ethylene leakage in complex environment.